\documentclass[preprint,12pt,authoryear]
{elsarticle}

\usepackage{amssymb}
\usepackage{longtable}
\usepackage{amsmath}
\usepackage[authoryear]{natbib} 
\usepackage{threeparttable}
\usepackage{multirow}
\usepackage{multicol}
\usepackage{graphicx}
\usepackage{subcaption}
\usepackage{xurl}
\usepackage{hyperref}
\usepackage[ruled,vlined]{algorithm2e}

\usepackage{algpseudocode}
\usepackage{longtable,booktabs,array}
\usepackage{amsthm}
\usepackage{tabularx,booktabs}

\begin{document}

\begin{frontmatter}

\title{Omni-scale Learning-based Sequential Decision Framework for Order Fulfillment of Tote-handling Robotic Systems} 

\author{Jiaxin Liu$^{a}$}

\author{Peng Yang$^{a,*}$}

\author{Yuping Li$^{a}$}

\author{Xinyue Xie$^{a}$}


\address{$^a$Institution of Data and Information, Shenzhen International Graduate School, Tsinghua University, Nanshan District, Shenzhen 518055, China}

\address{$^*$Corresponding author, email: yang.peng@sz.tsinghua.edu.cn}



\begin{abstract}
Driven by the rapid expansion of e-commerce and flexible, small-batch production, the size of the intralogistics load unit of finished goods, semi-finished goods and raw materials is steadily shrinking. Totes are gradually replacing pallets as the primary handling and storage container. This shift has propelled tote-handling robotic systems to the forefront of automation order fulfillment centers. The order-fulfillment decisions of tote-handling robotic systems share a common order-tote-robot sequential decision-making nature, that is independent of the specific system architecture. Existing studies primarily focus on decision mechanisms tailored to particular systems, making it difficult to generalize or transfer them to other contexts. We propose an Omni-scale Learning-based Sequential Decision Framework for Order Fulfillment of Tote-handling Robotic Systems (OLSF-TRS), a generalized and scalable sequential decision framework that combines structured combinatorial optimization with multi-agent reinforcement learning to coordinate order, tote, and robot decisions. On small-scale tote-handling robotic systems, OLSF-TRS achieves near-optimal performance with average optimality gaps below 3.5\% and millisecond-level inference times across two distinct system configurations. In large-scale, high-concurrency scenarios, OLSF-TRS consistently outperforms heuristic baselines across two different system types, reducing total tote movements by 8–12\% and over 30\% compared to state-of-the-art rule-based approaches, while maintaining real-time responsiveness. These improvements translate into tangible operational benefits, including cost reduction, lower energy consumption, and enhanced throughput stability—collectively advancing both economic efficiency and environmental sustainability. The proposed framework delivers an efficient and unified order fulfillment decision-making framework for widely deployed tote-handling robotic systems. Its modular and transferable design enables cost-effective adaptation across diverse system architectures, supporting high-quality order fulfillment in both e-commerce and industrial logistics sectors.
\end{abstract}




\end{frontmatter}


\section{Introduction}
\label{sec:intro}
Global supply chains are undergoing a profound transformation as the explosive growth of e-commerce and the rise of flexible, small-batch manufacturing place unprecedented pressure on intralogistics to fulfill massive small orders fast and accurately. Order volume increases while the order size gradually decreases over the past decade. The number of business-to-consumer parcels has more than tripled, while the average shipment weight has plummeted—UPS reports a 30\% drop in its domestic package weight between 2012 and 2022, and Cainiao estimates that more than 80\% of its 2023 Singles-Day parcels weighed less than 2 kg~\citep{UPS2012_10K,UPS2022_10K,UPU_PhysicalPostal2023}.

The demand for high-quality fulfillment of massive volumes of small orders has driven a continuous reduction in the size of standardized intralogistics handling containers. The demand for high-quality fulfillment of massive volumes of small orders has driven a continuous reduction in the size of standardized intralogistics handling containers, as finer-grained storage and retrieval units are required to efficiently process fragmented demand. As a result, totes are gradually replacing pallets as the primary handling and storage container in intralogistics, and tote-handling robotic systems have become the dominant form of automated order fulfillment centers. In recent years, a variety of innovative tote-handling robotic systems, such as Hairobotics, Skypod, and AirRob, have been widely deployed across e-commerce and industrial logistics scenarios~\citep{hairobotics2025cases,exotec2025skypod,AirRob}.

Tote-handling robotic systems usually feature a tote storage area, tote-handling robots, and tote-picking workstations. The order fulfillment process begins with incoming orders being assigned to specific slots on the put walls at the tote-picking workstations. Each slot holds one order box. An order box collects all items needed for an order. Totes containing the required items are identified and matched with the relevant orders stored in the tote storage area. Based on robots’ capacity, the matched totes are transferred into tote-handling tasks with the origin-destination information. These tasks are then distributed among the robots, which plan routes efficiently to deliver one or multiple totes to the workstations in a single trip. At the workstations, pickers retrieve items from the totes to complete orders. Once all items for an order are picked, the order box leaves the workstation, making space for new orders. This cycle continues until all orders are fulfilled.

The order-fulfillment decisions of tote-handling robotic systems share a common sequential decision-making process involving the order, tote, and robot, regardless of the specific system architecture. Differences among these systems primarily stem from the diverse features and structures of the tote-handling robots. For example, 2D Multi-Tote handling robots can transport multiple totes in a single trip, while 3D Rack-Climbing robots can climb shelves and move on the ground. The typical decision-making process includes: (i) order assignment, which determine how orders are allocated to tote-picking workstations and in what processing order; (ii) tote matching, which select the appropriate totes and determine their retrieval sequence according to order requirements; and (iii) robot scheduling, which assign robots to tote-handling tasks and schedule their operations over time. These insights motivate us to develop a generalized order fulfillment decision-making framework, which can serve as the software foundation for efficiently operating tote-handling robotic systems and support their rapid innovation and deployment.

Existing studies primarily focus on decision mechanisms tailored to particular systems, making it difficult to generalize or transfer them to other contexts.
A growing body of literature has investigated order-fulfillment decisions in multi-tote robotic warehouse systems, covering a broad spectrum of operational problems such as order assignment, tote retrieval, batching, and robot scheduling~\citep{qin2024making,qin2024performance,bai2025order,shan2024modeling}. From a structural perspective, these studies can be broadly categorized into three closely interrelated decision layers, namely order-level, tote-level, and robot-level decisions~\citep{deKoster2007, casella2023trends, aras2024order, boysen2019warehousing, vanheusden2023practical, shah2017comprehensive}.
Beyond algorithmic advances, recent work has also demonstrated that physical robotic systems can increasingly operate autonomously in cluttered, constrained environments and execute distributed transport coordination~\citep{Peng2025_NatCommun_AerialContinuum,Arbel2025_NatCommun_CooperativeTransport}, suggesting that the learning-based policies developed in this work are compatible with realistic deployment conditions. While this order-tote-robot layered decomposition provides useful analytical insights into the internal structure of fulfillment systems, most existing studies focus on optimizing individual layers in isolation. Such a fragmented approach limits the ability to capture the dynamic cross-layer coupling inherent in integrated fulfillment processes, as will be elaborated below.

Several studies have further explored task optimization and performance evaluation in specific tote-handling robotic warehouse systems, including rack-climbing robots, shuttle-based robotic storage systems, and other compact storage and retrieval technologies~\citep{bianco2025automated,de2022warehousing,arvind2025intelligent,trost2024analytical,mao2023research}. These systems differ substantially in their physical execution mechanisms: rack-climbing robots operate in three-dimensional shelf structures with vertical traversal costs that fundamentally shape routing decisions, while shuttle-based systems impose lane-constrained movement patterns that require dedicated sequencing logic, and multi-tote systems introduce concurrent load management constraints absent in single-tote designs. Consequently, the decision models and control policies developed for one system are often tightly coupled to its particular robotic platform, storage layout, and operational workflow, which restricts their applicability to broader warehouse contexts.

The integrated optimization of warehouse order fulfillment is inherently challenging because it constitutes a multi-stage joint combinatorial optimization problem in which each individual stage is
itself computationally hard, and the stages are strongly coupled with one another. Specifically, order batching and assignment constitutes a variant of bin-packing and scheduling; tote selection and sequencing is
a variant of the retrieval sequencing problem; and multi-robot task allocation and routing is a variant of the vehicle routing problem, all of which are NP-hard in general. Beyond the per-stage complexity, these
subproblems interact dynamically: decisions at the order level influence tote demand patterns and congestion dynamics, which in turn affect robot
routing and scheduling, while robot-level delays and resource contention feed back into order completion times. It is precisely this combination
of per-stage intractability and inter-stage coupling that renders the fulfillment problem resistant to conventional solution strategies, as each class of existing methods fails for a specific reason rooted in these structural properties.

Conventional rule-based and heuristic approaches~\citep{casella2023trends}
offer practical solutions but, by construction, optimize each layer under fixed rules that do not adapt to the dynamic inter-stage dependencies described above, making them unable to consistently account
for cross-stage coupling under varying operational conditions. Exact optimization methods, although theoretically sound, scale exponentially
with the combinatorial complexity of even a single stage; when extended to the joint multi-stage problem, they are limited to small-scale static
instances~\citep{Mohan_Banur_2024}, creating a widening gap between model-driven optimization and real-time operational requirements.

Building on these limitations, recent advances in artificial intelligence have opened new avenues for complex decision-making in warehouse logistics and operations research, particularly by enabling data-driven solution construction and adaptive control under uncertainty~\citep{sodiya2024ai,Begnardi2025sequential}. In particular, neural combinatorial optimization (NCO) has emerged as a promising paradigm that learns to construct high-quality solutions directly from problem instances, bypassing handcrafted heuristics~\citep{kool2019attention,drori2020learning}.
 NCO has been successfully applied to canonical problems such as the  Traveling Salesman Problem (TSP), Vehicle Routing Problem (VRP), Orienteering Problem (OP), and Knapsack Problem (KP)~\citep{nazari2018reinforcement,kool2019attention,vaswani2017attention,deudon2018learning,bello2017neural,bresson2021transformer}, highlighting its potential relevance to warehouse tasks with similar combinatorial structures. 
Complementarily, reinforcement learning (RL), particularly in large-scale and multi-agent settings, has demonstrated strong potential for coordinating complex networked systems under uncertainty. Recent advances in scalable MARL architectures have shown that near-optimal control policies can be learned efficiently even in massive, decentralized environments~\citep{yu2022surprising,ji2023multi,krnjaic2024scalable,ma2024efficient, xu2025multi,Su2025multiagent,Lu2025reward}. These methodological breakthroughs have significantly broadened the applicability of learning-based control to large robot fleets and logistics networks. Recent studies have further demonstrated the effectiveness of AI-driven approaches in integrated warehouse optimization tasks, including multi-stage order picking, storage assignment, and robot scheduling~\citep{monemi2025graph, liu2023integrated,ouhimmou2024machine,ahmad2025multi,zhang2025real,Bae2022Scientific}. 
Nevertheless, these works primarily highlight the growing applicability of AI and learning-based heuristics beyond isolated subproblems~\citep{qiu2025integrated,gioia2024rolling,hu2023dynamic}, yet they remain largely confined to specific simulation environments and often do not address the need for fully integrated decision pipelines.

Motivated by this, our goal is to provide a unifying, generalized framework that integrates learning-based methods with structured operational decision pipelines. 
Motivated by this, our goal is to provide a unifying, generalized framework that integrates learning-based methods with structured operational decision pipelines. The need for such a framework stems directly from two compounding gaps identified in the preceding review.
First, existing solutions are system-specific: decision models developed for a particular robotic architecture are tightly coupled to its platform, storage layout, and operational workflow, and cannot be transferred to a different system type without substantial re-engineering. Second, existing methods are scale-limited:
rule-based approaches fail to capture cross-stage dynamics under varying operational conditions, exact methods cannot scale beyond small static instances, and current learning-based works address isolated subproblems rather than the full sequential decision pipeline. A generalized framework addresses the first gap by abstracting heterogeneous system architectures into a common decision representation that supports diverse tote-handling platforms without platform-specific redesign. An omni-scale framework addresses the
second gap by operating effectively across the full spectrum of deployment sizes, from small-scale instances where exact solutions can be obtained and serve as supervision signals for policy learning,
to large-scale stochastic scenarios with hundreds of orders, dozens of robots, and tens of thousands of storage locations, where real-time coordination under high concurrency is required. Covering this full
scale spectrum is practically significant because deployed tote-handling systems span this entire range, and a unified framework eliminates the need to maintain separate decision pipelines for different deployment
scales.

We propose an Omni-scale Learning-based Sequential Decision Framework for Order Fulfillment of Tote-handling Robotic Systems (OLSF-TRS), which abstracts heterogeneous tote-handling robotic systems into a unified, data-driven sequential decision paradigm. At the methodological level, OLSF-TRS leverages bisimulation quotienting to construct abstract Markov decision processes (BQ-MDPs) to perform principled state space abstraction~\citep{drakulic2024bq,bai2016markovian,abel2018state}, enabling scalable learning across heterogeneous system configurations. 
Building upon this abstraction, bisimulation quotienting for neural combinatorial optimization (BQ-NCO) addresses the underlying combinatorial decision subproblems in a data-driven manner~\citep{kool2019attention}. 
To enhance scalability and robustness in large-scale stochastic environments, the resulting policy is optimized within a multi-agent proximal policy optimization (MAPPO) framework~\citep{schulman2017proximal}, allowing decentralized agents to learn coordinated policies under a shared global objective.

\begin{figure}[!ht]
    \centering
    \includegraphics[width=\linewidth]{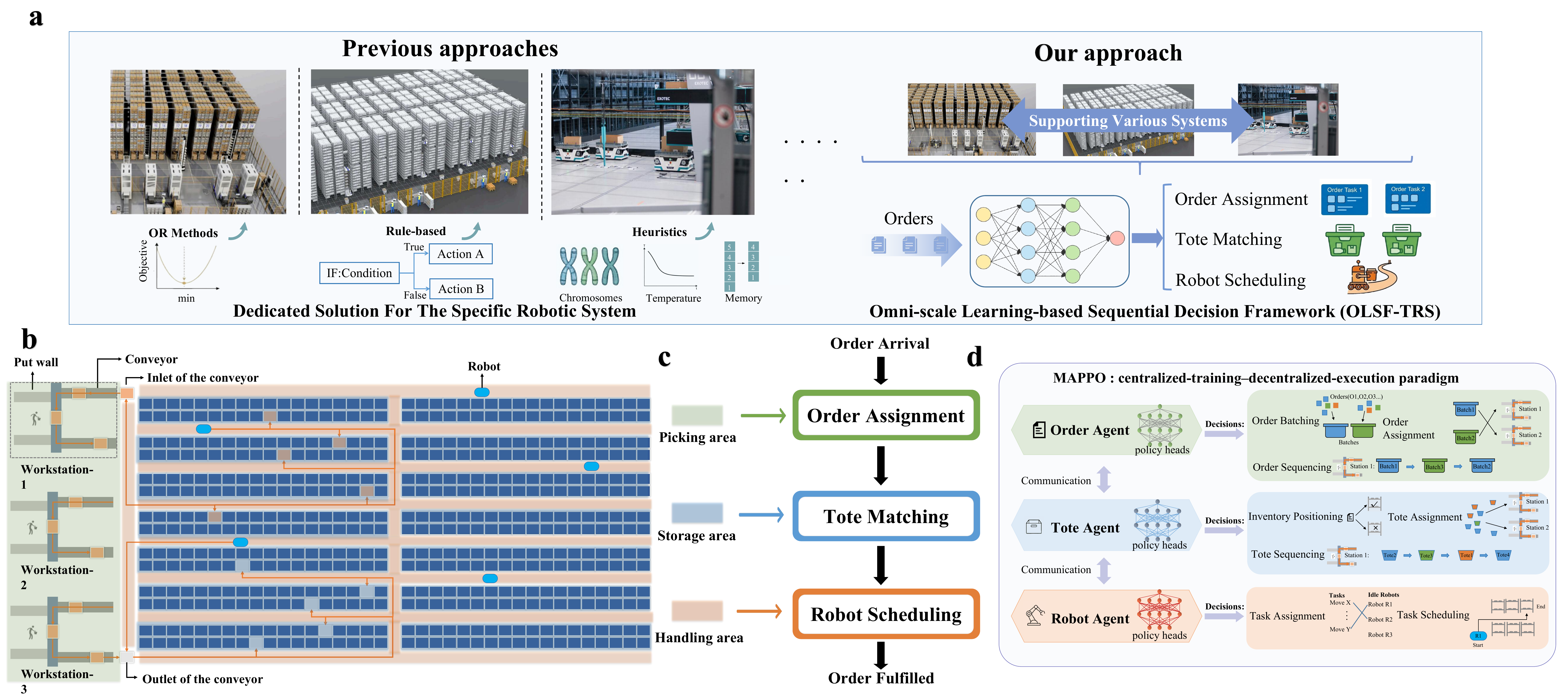}
    \caption{
    \textbf{Overview of the Omni-scale Learning-based Sequential Decision Framework for Order Ful-
fillment of Tote-handling Robotic Systems (OLSF-TRS).} 
    \textbf{a} Conceptual comparison between conventional system-specific fulfillment solutions and the proposed omni-scale learning-based framework. Traditional approaches rely on operations research methods, rule-based logic and heuristic pipelines that are typically tailored to a particular robotic architecture, whereas OLSF-TRS provides a unified learning framework that supports diverse tote-handling robotic systems and integrates order assignment, tote matching, and robot scheduling decisions within a common decision paradigm.
    \textbf{b} A simplified operational schematic of a tote-handling robotic fulfillment system, illustrating robot operations together with order sorting bays and conveyor structures.
    \textbf{c} A staged decision pipeline embedded in the warehouse operational context, where orders arriving to the system are sequentially processed through order assignment, tote matching, and robot scheduling decisions across picking, storage and handling areas until order fulfillment.
    \textbf{d} A hierarchical multi-agent decision architecture with decentralized execution, where Order, Tote, and Robot agents operate as cooperative decision units with inter-agent communication. Each agent is responsible for task-specific decisions, including order batching and sequencing, inventory positioning, tote sequencing and assignment, as well as robot task assignment and scheduling, enabling coordinated system-level fulfillment.}
    \label{fig:framework}
\end{figure}

Figure~\ref{fig:framework} provides an overview of the proposed OLSF-TRS. 
Figure~\ref{fig:framework}a contrasts conventional system-specific fulfillment solutions with the proposed generalized learning-based paradigm, highlighting the transition from rule-based, operations-research, and heuristic pipelines toward a unified and transferable decision framework that supports diverse tote-handling systems. 
Figure~\ref{fig:framework}b presents a simplified conceptual schematic of core operational interfaces in a robotic fulfillment setting, including the put wall, conveyor inlet and outlet, and robot interaction points. 
Figure~\ref{fig:framework}c illustrates a staged decision pipeline embedded in the warehouse operational context, where order arrival triggers sequential decisions of order assignment, tote matching, and robot scheduling across storage, picking, and handling areas until order fulfillment. 
Figure~\ref{fig:framework}d depicts a hierarchical decentralized multi-agent structure in which Order, Tote, and Robot agents act as cooperative decision units with communication links, each responsible for its corresponding decision tasks such as batching, assignment, sequencing, and scheduling.
Overall, the proposed OLSF-TRS  exhibits three key features:

\begin{enumerate}
    \item \textbf{A generalized sequential decision framework for order fulfillment of tote-handling robotic systems.}  
    We introduce a transferable decision-making framework that reframes order fulfillment as a structured sequence of interdependent decisions across order, tote, and robot levels. 
    By abstracting heterogeneous system architectures into a unified sequential representation, the framework provides a common conceptual and modeling foundation for analyzing, comparing, and controlling a wide class of robotic fulfillment systems.

    \item \textbf{A dependency-aware and modular decision paradigm.}  
    Unlike existing approaches that optimize isolated decision layers under fixed assumptions or tightly integrated formulations that couple multiple decisions into a single model, the proposed paradigm explicitly preserves cross-stage dependencies by linking all subproblems into a coherent decision pipeline. 
    This design enables modular training and deployment while maintaining system-level consistency, offering a principled alternative to both monolithic end-to-end optimization and fragmented heuristic control.

    \item \textbf{A scalable algorithmic realization via OLSF-TRS.}  
    We present a concrete learning-based realization that operationalizes the framework at scale. 
    OLSF-TRS integrates BQ-MDP for principled state abstraction, BQ-NCO for structured combinatorial decisions, and MAPPO for cooperative control in large-scale stochastic environments, enabling modular yet coordinated decision-making across order, tote, and robot levels. 
    This tightly coupled integration is specifically designed to support scalable deployment rather than serving as an isolated algorithmic contribution.
\end{enumerate}

\section{Results}\label{sec2}
\subsection{Overview of the OLSF-TRS }
The warehouse fulfillment environment comprises three primary entities, orders, totes, and robots, operating across multiple picking stations. Orders are characterized by SKU requirements, priority levels, and arrival sequences; totes serve as handling and storage containers with finite capacities; and robots execute tote-handling and transportation tasks. 

The proposed OLSF-TRS integrates structured neural combinatorial optimization with multi-agent reinforcement learning to enable coordinated, scalable decision-making across heterogeneous warehouse entities. Individual decision stages are equipped with BQ-based neural policy heads that efficiently approximate high-quality solutions for structured combinatorial subproblems, supporting adaptation to different fulfillment system architectures, including 2D Multi-Tote Handling Robotic Systems (Hairobotic) and 3D Rack-Climbing Robotic Systems (Exotec), as well as varying operational conditions such as different system scales, robot fleet sizes, and warehouse layouts.

During pretraining, BQ-NCO policy heads leverage expert trajectories generated from exact solvers and hybrid heuristics to capture near-optimal local decision strategies. The hierarchical integration of these pretrained policies within the MAPPO-based multi-agent learning framework ensures that local combinatorial decisions align with global operational goals, as evaluated through the system-level objective $Z_{\mathrm{Final}}$ (total tote handling events), reducing redundant tote movements and improving throughput under high concurrency.

This overview provides a conceptual foundation for the subsequent experimental evaluation, where OLSF-TRS is tested on both small-scale instances for pretraining validation and large-scale instances to assess system-level coordination and practical applicability in realistic warehouse scenarios.

\subsection{Data for Learning and Analysis}
The experimental instances are designed to emulate a practical automated tote-handling robotic warehouse. In real-world deployments, such systems typically consist of dense robotic storage areas composed of rack modules arranged in rows and columns.
A representative storage area may contain approximately 32 rack modules along the longitudinal direction and 4 modules along the transverse direction. Each rack module measures roughly 1.2 × 1.2 × 2.4 meters (length × width × height), consistent with standard shelving units, and contains multiple storage locations for tote storage. Across the entire storage area, this results in tens of thousands of storage locations, typically on the order of 50,000–75,000 locations in large-scale installations.
Each storage location accommodates a tote with a typical size of 0.6 × 0.4 × 0.3 meters (length × width × height). The system operates with 20–70 mobile robots and 8–25 picking or replenishment stations, serving dynamically arriving customer orders drawn from a SKU pool of several thousand products.

All instances are synthetically generated to emulate realistic tote-handling robotic warehouse systems. The parameter ranges, including SKU diversity, order volumes, robot quantities, workstation capacities, and tote availability, are randomly sampled 
within realistic operational ranges to represent typical operational variability observed in automated fulfillment centers. This controlled generation process ensures both experimental reproducibility and sufficient realism for system-level performance evaluation.

To evaluate the proposed OLSF-TRS, we construct a suite of 18 structured order fulfillment instance groups covering a spectrum from small, exactly solvable configurations to large-scale stochastic warehouse scenarios, as summarized in Table~\ref{tab:warehouse_instances}. Each instance specifies the number of SKUs, orders, robots, workstations, and totes, collectively defining the combinatorial complexity and coordination requirements of each decision stage. Totes are configured assuming a one-SKU-per-tote policy with redundant storage across the warehouse, and each workstation's capacity is defined by its associated put wall slots for concurrent order processing.

The first nine instances (S-1 to S-9) represent compact setups for supervised pretraining and validation of combinatorial policy heads. Optimal solutions for these instances are obtained using exact solvers, providing high-quality supervision for BQ-NCO policy heads and enabling verification of the optimality-preserving property of OLSF-TRS. Each subproblem includes approximately one thousand training instances and several hundred validation instances. The remaining nine instances (L-1 to L-15) correspond to large-scale
warehouse operations capturing realistic levels of stochasticity, resource contention, and agent concurrency, and are reserved for multi-agent reinforcement learning to assess OLSF-TRS's coordination
capability under high-density operations.

Experiments were conducted on different computing platforms depending on the problem scale. Supervised training and validation were performed on a workstation equipped with an Intel Core i7 processor, 32 GB RAM, and a 1 TB SSD, while an NVIDIA RTX 5090 GPU was used to accelerate neural network training. All implementations were developed in Python 3.11 using PyTorch 2.7.1 to enable efficient GPU-accelerated learning. The optimization baselines were solved using Gurobi Optimizer 12.0.0 through its official Python interface.

\begin{table}[!ht]

\centering

\caption{Configuration of warehouse instances used for evaluation. All instances specify SKUs, orders, robots, picking stations, and totes. Small-scale instances (S-1 to S-9) allow exact solution computation, while large-scale instances (L-1 to L-15) reflect realistic warehouse operations.}

\label{tab:warehouse_instances}

\begin{tabular}{cccccc}
\toprule
\textbf{Instance} & \textbf{SKUs} & \textbf{Orders} & \textbf{Robots} & \textbf{Workstations} & \textbf{Totes} \\
\midrule
S-1 & 20 & 10 & 3 & 3 & 100 \\
S-2 & 25 & 12 & 4 & 3 & 125 \\
S-3 & 30 & 14 & 4 & 4 & 150 \\
S-4 & 35 & 15 & 5 & 4 & 175 \\
S-5 & 40 & 15 & 5 & 5 & 200 \\
S-6 & 45 & 16 & 5 & 5 & 225 \\
S-7 & 50 & 18 & 6 & 5 & 250 \\
S-8 & 55 & 18 & 6 & 6 & 275 \\
S-9 & 60 & 20 & 6 & 6 & 300 \\
L-1 & 60 & 50 & 10 & 5 & 500 \\
L-2 & 80 & 60 & 15 & 5 & 650 \\
L-3 & 100 & 70 & 20 & 5 & 800 \\
L-4 & 120 & 80 & 25 & 10 & 1000 \\
L-5 & 140 & 90 & 30 & 10 & 1200 \\
L-6 & 160 & 100 & 35 & 10 & 1400 \\
L-7 & 180 & 110 & 40 & 15 & 1600 \\
L-8 & 200 & 120 & 45 & 15 & 1800 \\
L-9 & 220 & 130 & 50 & 15 & 2000 \\
L-10 & 250 & 150 & 55 & 15 & 2200 \\
L-11 & 300 & 180 & 60 & 20 & 2500 \\
L-12 & 350 & 210 & 60 & 20 & 3000 \\
L-13 & 400 & 250 & 65 & 20 & 3500 \\
L-14 & 500 & 300 & 65 & 20 & 4000 \\
L-15 & 600 & 350 & 70 & 25 & 5000 \\
\bottomrule
\end{tabular}
\end{table}

\subsection{Performance on Small-Scale Warehouse Instances}

We first evaluate the proposed OLSF-TRS on small-scale warehouse instances to verify the accuracy of the learned policies at the individual decision unit level. At this scale, the primary objective is to ensure that the imitation learning policy produces high-quality allocations and sequences, closely approximating the optimal solutions obtained by exact solvers. This modular validation guarantees that the subsequent multi-agent coordination in larger systems is built upon reliable local decision-making.

For each agent, several hundred test instances are used to obtain stable performance estimates without incurring excessive computational overhead. Baseline solutions were obtained using the Gurobi exact solver, providing a gold standard for optimality. Performance metrics include Average Gap (\%). Here, the objective value refers to the total number of tote movements, including both retrievals and the subsequent restorage (returns), required to fulfill all orders in a given instance. The Average Gap is calculated relative to the baseline objectives as:
\begin{equation}
\label{eqgap_modular}
\text{Average Gap} = \frac{1}{N} \sum_{i=1}^N \frac{ f_i^\text{OLSF-TRS}-f_i^\text{baseline}}{f_i^\text{baseline}} \times 100\%,
\end{equation}
where $f_i^\text{OLSF-TRS}$ denotes the total number of tote movements returned by the learned policy for instance $i$.

\begin{table}[!ht]
\centering
\caption{Agent-level performance of OLSF-TRS on small-scale warehouse instances (S-1 to S-9). The Gap measures the percentage of additional tote movements (including retrievals and restorage) relative to the Gurobi optimal solution.}
\label{tab:small_scale_ilwff_modules_compact}
\scriptsize
\begin{tabular}{lcccccc}
\toprule
\textbf{Instance} 
& \multicolumn{2}{c}{\textbf{Order}} 
& \multicolumn{2}{c}{\textbf{Tote}} 
& \multicolumn{2}{c}{\textbf{Robot}} \\
\cmidrule(lr){2-3} \cmidrule(lr){4-5} \cmidrule(lr){6-7}
& Gap (\%) & Time (ms) & Gap (\%) & Time (ms) & Gap (\%) & Time (ms) \\
\midrule
S-1 & 3.0 & 38.2 & 3.3 & 42.5 & 3.5 & 51.0 \\
S-2 & 3.1 & 39.0 & 3.4 & 43.0 & 3.6 & 52.0 \\
S-3 & 3.2 & 40.0 & 3.5 & 44.0 & 3.7 & 53.0 \\
S-4 & 3.3 & 41.0 & 3.6 & 45.0 & 3.8 & 54.0 \\
S-5 & 3.4 & 42.0 & 3.7 & 46.0 & 3.9 & 55.0 \\
S-6 & 3.5 & 43.0 & 3.8 & 47.0 & 4.0 & 56.0 \\
S-7 & 3.6 & 44.0 & 3.9 & 48.0 & 4.1 & 57.0 \\
S-8 & 3.7 & 45.0 & 4.0 & 49.0 & 4.2 & 58.0 \\
S-9 & 3.8 & 46.0 & 4.1 & 50.0 & 4.3 & 59.0 \\
\bottomrule
\end{tabular}
\end{table}

As shown in Table~\ref{tab:small_scale_ilwff_modules_compact}, the OLSF-TRS achieves near-optimal performance across all agents, with average gaps below 3.5\%. The Order agent, with relatively compact state and action spaces, demonstrates the lowest gap, while the Robot agent, due to the combinatorial complexity of task allocation, exhibits slightly higher gaps. Success rates remain above 95\%, and inference times are within tens of milliseconds, confirming that the learned policies are both accurate and computationally efficient. These results validate that the OLSF-TRS can reliably produce high-quality local decisions, providing a solid foundation for subsequent system-level multi-agent coordination experiments.

\subsection{Performance of OLSF-TRS on Large-Scale Warehouse Instances}
Building on the validated performance of OLSF-TRS on small-scale instances, we next evaluate the framework in large-scale warehouse scenarios that more closely resemble practical automated fulfillment environments. These scenarios introduce both stochasticity and concurrency inherent in real-world operations. Stochasticity is reflected in randomly sampled order arrival times, variable SKU combinations within each order, and heterogeneous initial placements of robots and totes. Concurrency arises from multiple robots simultaneously executing tote retrieval and delivery tasks, as well as multiple human-operated workstations concurrently processing incoming totes. Together, these characteristics create a dynamic and densely interacting system, challenging the decision-making framework to maintain high-quality, coordinated performance.

Exact optimization is computationally intractable at this scale, making these experiments a test of OLSF-TRS's generalization and practical applicability. Each large-scale instance is evaluated over several hundred simulation runs to account for stochastic variability, providing stable performance estimates while controlling computational overhead. Within the framework, BQ-NCO policy heads provide high-quality, structured decision policies for individual agents (orders, totes, and robots), while the centralized MAPPO critic enables these heterogeneous agents to coordinate effectively, capturing inter-agent dependencies and mitigating conflicts under high-density, concurrent operations.

Tables~\ref{tab:hai_runtime_s} and \ref{tab:skypod_runtime_s} report the raw quantitative performance of OLSF-TRS and baseline methods in terms of total tote moves and cumulative runtime under the 2D Multi-Tote Handling Robotic Systems (Hairobotic) and 3D Rack-Climbing Robotic Systems (Exotec), respectively.

\begin{figure*}[!ht]
    \centering
    \includegraphics[width=\textwidth]{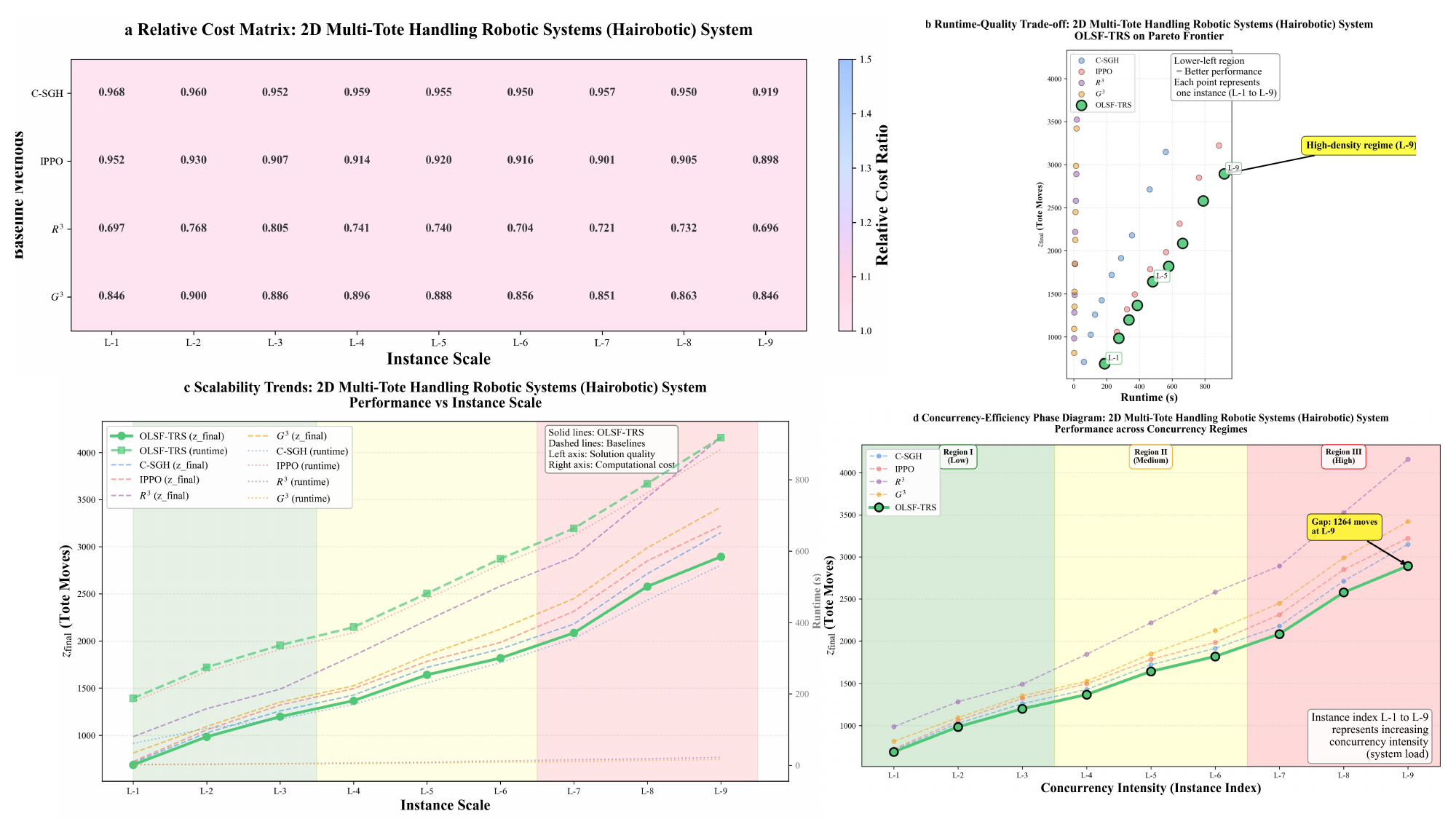}
    \caption{
    \textbf{System-level performance analysis of OLSF-TRS under the 2D Multi-Tote Handling Robotic Systems (Hairobotic).}
     \textbf{a} Relative tote-move cost matrix comparing OLSF-TRS with baseline methods across large-scale instances (L-1 to L-9), where each entry denotes the normalized cost ratio with respect to the corresponding baseline.
    \textbf{b} Runtime–quality trade-off illustrating the Pareto efficiency of OLSF-TRS in terms of total tote moves ($z_{\text{final}}$) and cumulative simulation runtime.
    \textbf{c} Scalability trends of operational cost and simulation runtime as a function of increasing system load, highlighting the stability of OLSF-TRS under growing concurrency.
    \textbf{d} Concurrency–efficiency phase diagram depicting the relationship between agent density and normalized tote-move cost, emphasizing the robustness of OLSF-TRS in high-density regimes.
    }
    \label{fig:system_level_hai}
\end{figure*}

\begin{figure*}[!ht]
    \centering
    \includegraphics[width=\textwidth]{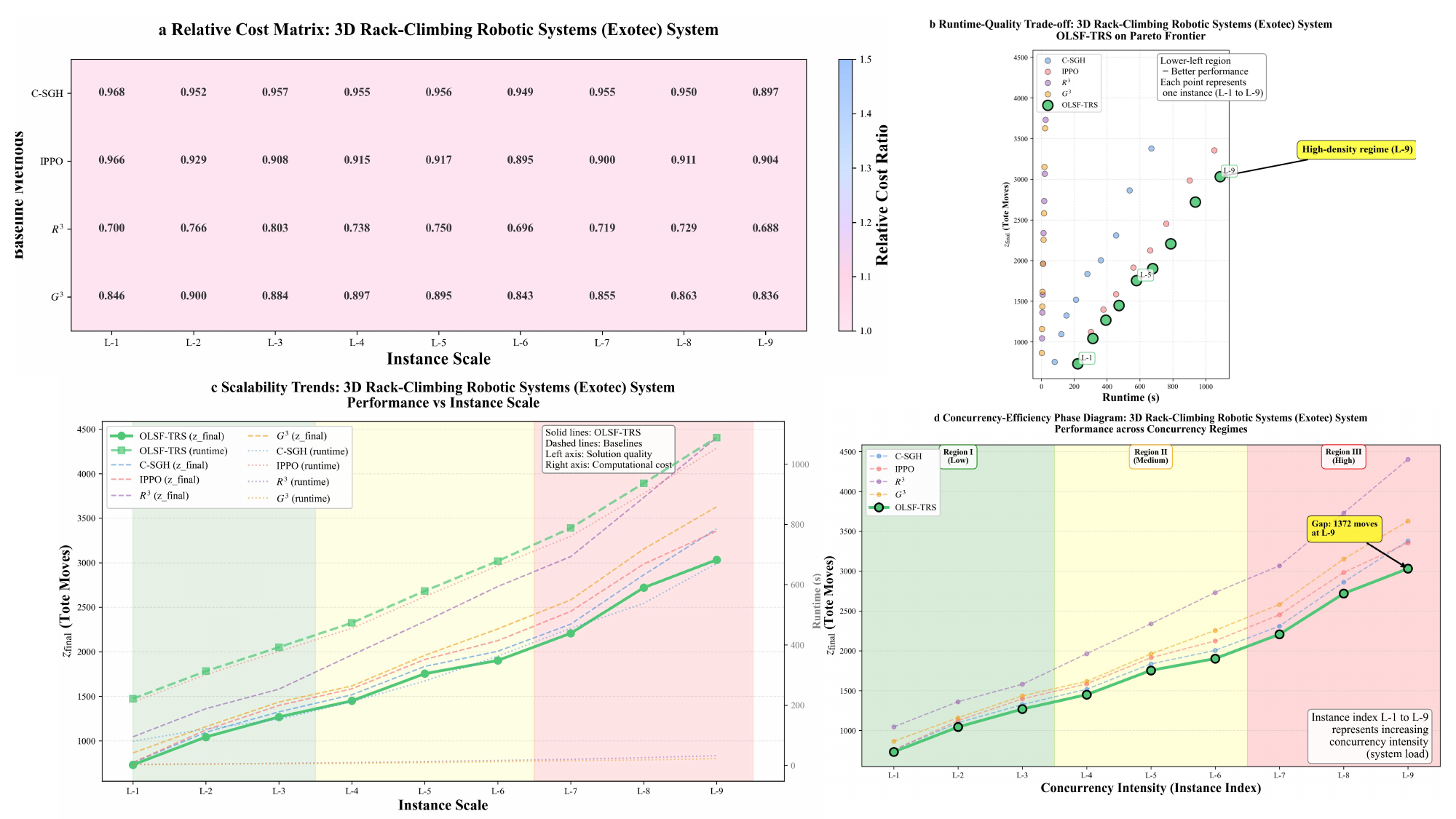}
    \caption{
    \textbf{System-level performance analysis of OLSF-TRS under the 3D Rack-Climbing Robotic Systems (Exotec).}
 \textbf{a} Relative tote-move cost matrix comparing OLSF-TRS with baseline methods across large-scale instances (L-1 to L-9), where each entry denotes the normalized cost ratio with respect to the corresponding baseline.
   \textbf{b} Runtime–quality trade-off illustrating the Pareto efficiency of OLSF-TRS in terms of total tote moves ($z_{\text{final}}$) and cumulative simulation runtime.
  \textbf{c} Scalability trends of operational cost and simulation runtime as a function of increasing system load, highlighting the stability of OLSF-TRS under growing concurrency.
 \textbf{d} Concurrency–efficiency phase diagram depicting the relationship between agent density and normalized tote-move cost, emphasizing the robustness of OLSF-TRS in high-density regimes.
    }
    \label{fig:system_level_skypod}
\end{figure*}

To further elucidate OLSF-TRS performance trends beyond tabulated results, Fig.~\ref{fig:system_level_hai} and Fig.~\ref{fig:system_level_skypod} present complementary visual diagnostics for the 2D Multi-Tote Handling Robotic Systems (Hairobotic) and 3D Rack-Climbing Robotic Systems (Exotec), respectively. Fig.~\ref{fig:system_level_hai}(a) and Fig.~\ref{fig:system_level_skypod}(a) report relative tote-move cost matrices, where each entry denotes the normalized ratio between OLSF-TRS and a given baseline under the same instance. Across both systems, OLSF-TRS consistently yields cost ratios below unity, with the most pronounced gains observed in high-density scenarios (L-7 to L-9). This pattern confirms that the performance advantage of OLSF-TRS is not confined to specific instance scales but instead amplifies as system concurrency increases.

The runtime–quality trade-offs are illustrated in Fig.~\ref{fig:system_level_hai}(b) and Fig.~\ref{fig:system_level_skypod}(b), where each method is positioned in the two-dimensional space spanned by cumulative runtime and total tote moves. OLSF-TRS forms a clear Pareto-efficient frontier, achieving substantially lower operational cost than heuristic baselines while incurring only moderate increases in simulation time. In contrast, C-SGH exhibits rapidly growing runtime without commensurate quality improvements, and decentralized Independent Proximal Policy Optimization (IPPO) drifts toward inferior trade-off regions under heavy system load.

Scalability trends are further visualized in Fig.~\ref{fig:system_level_hai}(c) and Fig.~\ref{fig:system_level_skypod}(c). As the system load increases from L-1 to L-9, OLSF-TRS maintains a near-linear growth profile in tote moves, whereas heuristic baselines display superlinear escalation, particularly in the 3D Rack-Climbing Robotic Systems. Notably, the runtime growth of OLSF-TRS is dominated by the high-fidelity simulator rather than by policy inference, which remains below 10 ms per decision step across all instances.

Finally, the concurrency–efficiency phase diagrams in Fig.~\ref{fig:system_level_hai}(d) and Fig.~\ref{fig:system_level_skypod}\ (d) highlight the robustness of OLSF-TRS under dense multi-agent interactions. As agent density increases, decentralized methods such as IPPO experience marked efficiency degradation due to confounded reward signals and uncoordinated congestion effects. In contrast, OLSF-TRS exhibits a significantly flatter efficiency decay curve, indicating that the integration of structured combinatorial policies with centralized multi-agent coordination effectively mitigates congestion-induced inefficiencies.

Taken together, these visual diagnostics corroborate the quantitative results in Tables~\ref{tab:hai_runtime_s} and \ref{tab:skypod_runtime_s}, demonstrating that OLSF-TRS delivers consistent system-level advantages across heterogeneous warehouse architectures. The observed performance gains stem not only from improved local decision quality but also from emergent global coordination effects, which become increasingly critical as warehouse scale and concurrency intensify.

\begin{table*}[!ht]
\centering
\caption{Comparison of OLSF-TRS and Baseline Methods under the 2D Multi-Tote Handling Robotic Systems in Terms of Tote Moves and Runtime (s).}
\label{tab:hai_runtime_s}
\resizebox{\textwidth}{!}{
\begin{tabular}{lcccccccccc}
\toprule
\multirow{2}{*}{Instance} 
& \multicolumn{5}{c}{Tote Moves ($z_{\text{final}}$)} 
& \multicolumn{5}{c}{Runtime (s)} \\
\cmidrule(lr){2-6} \cmidrule(lr){7-11}
& OLSF-TRS & C-SGH & IPPO & $R^3$ & $G^3$
& OLSF-TRS & C-SGH & IPPO & $R^3$ & $G^3$ \\
\midrule
L-1 & 687 & 710 & 722 & 985 & 812 & 188.1 & 61.3 & 178.1 & 2.1 & 1.5 \\
L-2 & 984 & 1025 & 1058 & 1282 & 1093 & 274.6 & 103.2 & 262.7 & 3.4 & 2.2 \\
L-3 & 1198 & 1258 & 1321 & 1489 & 1352 & 336.4 & 129.6 & 324.6 & 5.1 & 3.8 \\
L-4 & 1367 & 1426 & 1495 & 1845 & 1525 & 387.1 & 170.5 & 371.3 & 6.8 & 4.5 \\
L-5 & 1642 & 1720 & 1784 & 2219 & 1850 & 481.4 & 231.1 & 466.0 & 8.5 & 6.1 \\
L-6 & 1819 & 1915 & 1985 & 2582 & 2126 & 578.7 & 287.6 & 562.7 & 12.4 & 8.9 \\
L-7 & 2086 & 2180 & 2315 & 2892 & 2450 & 663.5 & 355.2 & 645.1 & 15.6 & 10.2 \\
L-8 & 2579 & 2714 & 2850 & 3525 & 2989 & 789.2 & 462.6 & 763.1 & 18.2 & 13.5 \\
L-9 & 2894 & 3150 & 3223 & 4158 & 3420 & 918.1 & 560.2 & 885.2 & 22.5 & 16.8 \\

\bottomrule
\end{tabular}
}
\end{table*}
\begin{table*}[!ht]
\centering
\caption{Comparison of OLSF-TRS and Baseline Methods under the 2D Multi-Tote Handling Robotic Systems (Hairobotic) on Extended Large-Scale Instances in Terms of Tote Moves and Runtime (s).}
\label{tab:hai_runtime_s_ext}
\resizebox{\textwidth}{!}{
\begin{tabular}{lcccccccccc}
\toprule
\multirow{2}{*}{Instance} 
& \multicolumn{5}{c}{Tote Moves ($z_{\text{final}}$)} 
& \multicolumn{5}{c}{Runtime (s)} \\
\cmidrule(lr){2-6} \cmidrule(lr){7-11}
& OLSF-TRS & C-SGH & IPPO & $R^3$ & $G^3$
& OLSF-TRS & C-SGH & IPPO & $R^3$ & $G^3$ \\
\midrule
L-10 & 3313 & 3621 & 3712 & 4823 & 3950 & 1065.4 & 658.4 & 1025.6 & 26.8 & 19.8 \\
L-11 & 3784 & 4182 & 4310 & 5652 & 4581 & 1228.6 & 782.5 & 1185.2 & 32.5 & 24.2 \\
L-12 & 4325 & 4855 & 5026 & 6685 & 5354 & 1415.2 & 925.8 & 1368.4 & 39.2 & 29.5 \\
L-13 & 5052 & 5752 & 5987 & 8053 & 6382 & 1672.8 & 1128.6 & 1620.5 & 48.6 & 36.8 \\
L-14 & 6124 & 7083 & 7423 & 10155 & 7854 & 2052.5 & 1425.2 & 1992.8 & 62.4 & 47.2 \\
L-15 & 7280 & 8566 & 9058 & 12584 & 9527 & 2486.3 & 1782.4 & 2418.6 & 78.5 & 59.6 \\
\bottomrule
\end{tabular}
}
\end{table*}
\begin{table*}[!ht]
\centering
\caption{Comparison of OLSF-TRS and Baseline Methods under the 3D Rack-Climbing Robotic Systems (Exotec) on Extended Large-Scale Instances in Terms of Tote Moves and Runtime (s).}
\label{tab:skypod_runtime_s_ext}
\resizebox{\textwidth}{!}{
\begin{tabular}{lcccccccccc}
\toprule
\multirow{2}{*}{Instance} 
& \multicolumn{5}{c}{Tote Moves ($z_{\text{final}}$)} 
& \multicolumn{5}{c}{Runtime (s)} \\
\cmidrule(lr){2-6} \cmidrule(lr){7-11}
& OLSF-TRS & C-SGH & IPPO & $R^3$ & $G^3$
& OLSF-TRS & C-SGH & IPPO & $R^3$ & $G^3$ \\
\midrule
L-10 & 3484 & 3890 & 3881 & 5127 & 4183 & 5082.4 & 3185.6 & 4920.8 & 38.5 & 26.8 \\
L-11 & 3985 & 4522 & 4527 & 6024 & 4877 & 5920.6 & 3782.4 & 5742.5 & 47.2 & 32.5 \\
L-12 & 4560 & 5283 & 5310 & 7152 & 5728 & 6885.2 & 4512.8 & 6685.2 & 57.8 & 39.8 \\
L-13 & 5340 & 6325 & 6382 & 8689 & 6853 & 8245.8 & 5528.6 & 8015.6 & 72.4 & 50.2 \\
L-14 & 6480 & 7858 & 7962 & 10989 & 8483 & 10250.4 & 7052.4 & 9985.4 & 94.5 & 65.4 \\
L-15 & 7727 & 9569 & 9781 & 13656 & 10354 & 12586.2 & 8852.5 & 12285.6 & 120.8 & 83.6 \\
\bottomrule
\end{tabular}
}
\end{table*}

To further assess the scalability of OLSF-TRS under increasingly resource-constrained conditions, we extend the evaluation to six additional large-scale instances (L-10 to L-15), as detailed in Table~\ref{tab:warehouse_instances}. These instances progressively increase the number of SKUs (from 250 to 600), orders (from 150 to 350), and totes (from 2{,}200 to 5{,}000), while the robot fleet grows only moderately from 55 to 70 units. This deliberate design creates a widening gap between task demand and robot availability: the robot-to-order ratio decreases from 0.37 in L-10 to 0.20 in L-15, imposing progressively tighter resource bottlenecks and intensifying scheduling contention among agents.

Tables~\ref{tab:hai_runtime_s_ext} and \ref{tab:skypod_runtime_s_ext} report the quantitative results of OLSF-TRS and baseline methods on these extended instances under the 2D Multi-Tote Handling Robotic Systems (Hairobotic) and 3D Rack-Climbing Robotic Systems (Exotec), respectively. Across both system architectures, OLSF-TRS consistently achieves the lowest tote-move count among all methods. Under the Hairobotic system, OLSF-TRS reduces tote moves by approximately 8.6\% relative to C-SGH and 10.9\% relative to IPPO at the L-10 scale; these margins widen to 14.9\% and 19.6\%, respectively, at the L-15 scale. Against the rule-based baselines $R^3$ and $G^3$, the improvements are even more substantial, with OLSF-TRS achieving 42.1\% and 23.5\% fewer tote moves at L-15 under the Hairobotic system. Similar trends are observed under the Exotec system, where OLSF-TRS attains reductions of up to 43.4\% relative to $R^3$ and 25.4\% relative to $G^3$ at L-15.

Notably, under the resource-constrained regime of L-14 and L-15, heuristic baselines such as $R^3$ and $G^3$ exhibit substantial performance degradation, as their rule-based strategies lack the capacity to resolve complex scheduling conflicts arising from robot scarcity. Decentralized learning methods such as IPPO also suffer from confounded reward signals and uncoordinated congestion, leading to superlinear growth in tote moves. In contrast, OLSF-TRS benefits from the MAPPO-based centralized critic, which enables effective global coordination even when individual agents face tight capacity constraints. The runtime of OLSF-TRS scales moderately with instance size, with the majority of computational cost attributable to the high-fidelity simulation environment rather than policy inference, which remains below 15\,ms per decision step across all extended instances.

These extended experiments demonstrate that OLSF-TRS not only scales effectively to warehouse configurations significantly larger than those in the initial evaluation, but also exhibits robust performance under conditions of increasing resource scarcity, a critical property for practical deployment in automated fulfillment centers where robot fleets are typically sized conservatively relative to peak order volumes.

\begin{table*}[!ht]
\centering
\caption{Comparison of OLSF-TRS and Baseline Methods under the 3D Rack-Climbing Robotic Systems (Exotec) in Terms of Tote Moves and Runtime (s).}
\label{tab:skypod_runtime_s}
\resizebox{\textwidth}{!}{
\begin{tabular}{lcccccccccc}
\toprule
\multirow{2}{*}{Instance} 
& \multicolumn{5}{c}{Tote Moves ($z_{\text{final}}$)} 
& \multicolumn{5}{c}{Runtime (s)} \\
\cmidrule(lr){2-6} \cmidrule(lr){7-11}
& OLSF-TRS & C-SGH & IPPO & $R^3$ & $G^3$
& OLSF-TRS & C-SGH & IPPO & $R^3$ & $G^3$ \\
\midrule
L-1 & 731 & 755 & 757 & 1045 & 864 & 885.2 & 320.4 & 840.1 & 3.5 & 2.1 \\
L-2 & 1043 & 1096 & 1123 & 1361 & 1159 & 1250.6 & 485.2 & 1205.4 & 5.1 & 3.8 \\
L-3 & 1268 & 1325 & 1397 & 1580 & 1435 & 1568.4 & 610.8 & 1512.6 & 7.2 & 5.4 \\
L-4 & 1449 & 1517 & 1584 & 1963 & 1616 & 1892.1 & 845.3 & 1820.7 & 9.4 & 6.8 \\
L-5 & 1754 & 1835 & 1913 & 2340 & 1960 & 2315.7 & 1120.6 & 2240.8 & 12.8 & 9.2 \\
L-6 & 1902 & 2005 & 2124 & 2732 & 2256 & 2712.4 & 1450.2 & 2650.3 & 16.5 & 12.4 \\
L-7 & 2207 & 2310 & 2453 & 3068 & 2582 & 3150.9 & 1820.5 & 3040.6 & 20.4 & 15.6 \\
L-8 & 2719 & 2863 & 2984 & 3731 & 3152 & 3745.2 & 2150.8 & 3610.4 & 25.8 & 18.2 \\
L-9 & 3032 & 3380 & 3355 & 4404 & 3628 & 4350.8 & 2680.4 & 4210.5 & 32.4 & 22.5 \\
\bottomrule
\end{tabular}
}
\end{table*}

\section{Methods}

\subsection{Problem Modeling as BQ-MDPs}
\label{bqmdp}

Each decision stage in the order fulfillment process of the tote-handling robotic system is first formulated as a Markov Decision Process (MDP), defined by the tuple $(\mathcal{S}, \mathcal{A}, P, R, \gamma)$, where $\mathcal{S}$ denotes the state space, $\mathcal{A}$ the action space, $P(s'|s,a)$ the state transition probability function, $R(s,a)$ the reward function, and $\gamma \in [0,1]$ the discount factor. 

To mitigate the exponential growth of the state space in combinatorial optimization, bisimulation quotienting is applied to construct an abstracted MDP, referred to as a BQ-MDP. In the BQ-MDP, original states $s_1, s_2 \in \mathcal{S}$ are grouped into abstract states $\tilde{\mathcal{S}}$ if they are bisimulation equivalent, i.e., for all actions $a \in \mathcal{A}$, they produce identical transition distributions $P(s'|s_1,a) = P(s'|s_2,a)$ and expected rewards $R(s_1,a) = R(s_2,a)$.  The condition $R(s_1,a) = R(s_2,a)$ holds when different physical configurations yield the same operational impact. For example, two robots at mirrored positions relative to a workstation will incur identical energy costs (negative rewards) to reach that target. Similarly, two different totes containing the same SKU requested by an order provide the same fulfillment reward when delivered. 
This abstraction substantially reduces the state-space dimensionality because it collapses the vast number of redundant configurations in large-scale warehouses (e.g., 50,000+ storage cells) into a compact set of behaviorally equivalent classes. Specifically, states are considered equivalent if they share identical decision-relevant attributes, such as homogeneous robots equidistant to a task or storage locations with identical SKU profiles and travel-time costs to workstations. This process ensures the optimal policy structure is preserved while keeping the problem computationally tractabl.

Based on the resulting abstract states, the BQ-NCO module learns stage-specific neural policies that serve as the core decision engines within the sequential framework. The dataset of abstract state--action pairs $(s,a)$ captures behaviorally relevant information for policy learning.

\subsection{OLSF-TRS Algorithm}
\begin{figure}[!ht]
\centering
\includegraphics[width=\linewidth]{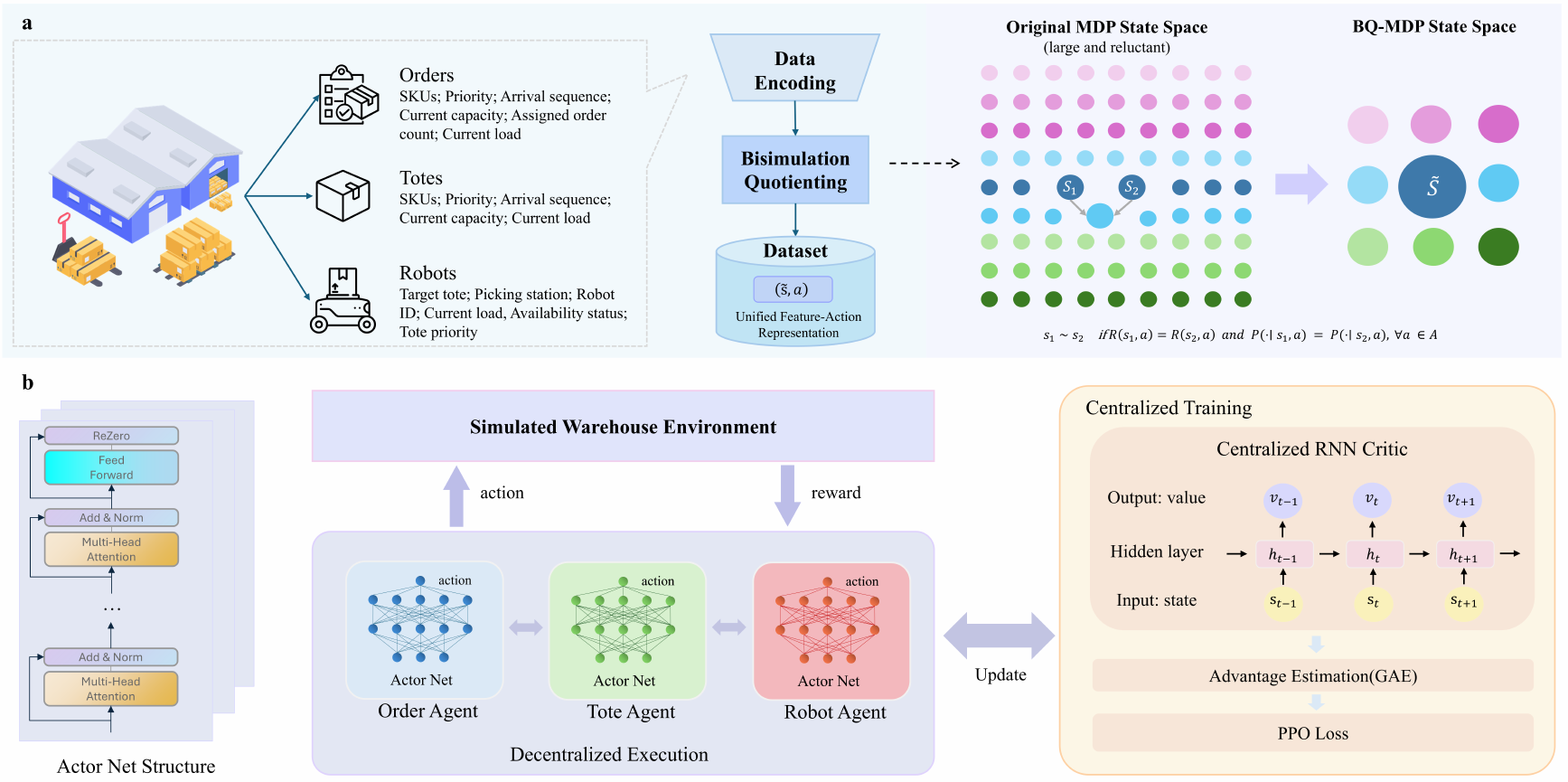}
\caption{
\textbf{Architecture of the OLSF-TRS for tote-handling robotic fulfillment.}
\textbf{a} Heterogeneous warehouse entities, including orders, totes, and robots, are represented by their relevant attributes such as SKUs, priority, arrival sequence, capacities, and current loads. These entities are encoded into a unified feature–action representation. The original large-scale MDP state space is then abstracted into a reduced BQ-MDP state space via bisimulation quotienting, where states with equivalent reward and transition dynamics are grouped together. The resulting dataset of state–action pairs $(s, a)$ captures the behaviorally relevant information for subsequent policy learning.
\textbf{b} Multi-agent reinforcement learning architecture for large-scale stochastic environments. Order, Tote, and Robot agents are implemented as attention-enhanced actor networks interacting with a simulated warehouse environment. Training follows a centralized-training–decentralized-execution paradigm, where a centralized recurrent critic evaluates joint states and actions, and policy updates are performed using advantage estimation and PPO-based optimization.
\label{fig:ilwff_architecture}
}
\end{figure}

Figure~\ref{fig:ilwff_architecture} presents the OLSF-TRS architecture, illustrating how heterogeneous warehouse entities are encoded into a unified feature--action representation, abstracted via bisimulation quotienting, and integrated with BQ-NCO pretraining and MAPPO-based multi-agent coordination for scalable decision making.

\subsubsection{Unified Feature--Action Representation and Neural Policy Learning}
All decision modules adopt a unified feature--action representation to facilitate parameter sharing and cross-stage generalization. Continuous attributes are standardized, while categorical attributes are encoded using learnable embeddings. Action spaces are defined by operational constraints, and feasibility masks enforce valid decisions. Table~\ref{tab:feature_action_module} summarizes the feature–action design for orders, totes, and robots.

\begin{table}[!ht]
\centering
\caption{Unified Feature--Action Design for Warehouse Order Fulfillment Modules}
\begin{tabular}{p{0.15\linewidth}<{\centering}p{0.25\linewidth}<{\raggedright}p{0.50\linewidth}<{\raggedright}}  
\toprule
\textbf{Module} & \textbf{Typical Decisions} & \textbf{Key Features / Action Space} \\
\midrule
Order  & Batching, Assignment, Sequencing & Order SKUs, priority, arrival sequence, workstation/batch capacity; actions: assign order to batch/picking station, determine order sequence \\
Tote   & Allocation, Sequencing           & Tote capacity, current load, order requirements; actions: assign tote to order, determine tote processing sequence \\
Robot  & Task Allocation, Scheduling      & Robot availability, target tote, picking station; actions: assign task to robot, determine temporal execution order \\
\bottomrule
\end{tabular}
\label{tab:feature_action_module}
\end{table}

For each BQ-MDP stage $k \in \{1, \dots, K\}$, a stage-specific policy $\pi_k(a|s)$ is learned using a shared Transformer encoder. The input feature matrix $X \in \mathbb{R}^{n \times d_\mathrm{in}}$ represents $n$ entities (orders, totes, or robots), each with $d_\mathrm{in}$ features, is processed by a fully connected linear unit. This linear unit maps the heterogeneous input features into a 128-dimensional latent space, followed by a Rectified Linear Unit (ReLU) activation to capture non-linear characteristics:
\begin{equation}
H^{(0)} = \mathrm{ReLU}(X W_0 + b_0),
\end{equation}
where $W_0$ and $b_0$ are the learnable weight matrix and bias vector of the linear unit, respectively. $H^{(0)}$ is processed by a 3-layer, 12-head Transformer encoder with ReZero normalization. The output is linearly projected to action logits $z_a$ for each candidate action $a \in \mathcal{A}_k$, with infeasible actions masked in the valid action set $\mathcal{A}_{k,\mathrm{valid}}$:
\begin{equation}
\pi_k(a|s) = \frac{\exp(z_a)}{\sum_{a' \in \mathcal{A}_{k,\mathrm{valid}}} \exp(z_{a'})}.
\end{equation}

Model training is conducted via imitation learning using expert trajectories obtained from exact solvers (e.g., Gurobi) on small-scale instances (S-1 to S-9). Each trajectory consists of a sequence of state–action pairs $(s_t, a_t)$, where $s_t$ represents the warehouse state at decision step $t$ and $a_t$ the corresponding optimal action. To improve generalization across different instance configurations and mitigate overfitting, expert trajectories are decomposed into subsequences of varying lengths, effectively augmenting the training dataset and exposing the policy to a broader variety of combinatorial patterns. This procedure allows the policy to capture both local and long-horizon dependencies inherent in sequential warehouse decision-making.

Because the framework operates as a sequential decision pipeline, errors made in upstream stages can propagate downstream, potentially compounding suboptimal decisions. To mitigate this effect, a soft coupling mechanism is employed: the predicted action vector $I_{k+1}^{\mathrm{pred}}$ from stage $k$ is aligned with the ground-truth action vector $I_{k+1}^{\mathrm{true}}$ used as input for stage $k+1$. The training loss at stage $k$ is therefore defined as
\begin{equation}
\mathcal{L}k = \mathcal{L}k^{\mathrm{CE}} + \lambda_k \left| I{k+1}^{\mathrm{pred}} - I{k+1}^{\mathrm{true}} \right|_2^2,
\end{equation}
where $\mathcal{L}_k^{\mathrm{CE}}$ is the standard cross-entropy loss for the stage-specific decision, $\lambda_k$ is a tunable weight controlling the strength of coupling between consecutive stages, and $|\cdot|_2$ denotes the L2 norm. The first term ensures that the policy reproduces expert actions at the current stage, while the second term penalizes discrepancies that could propagate to subsequent stages, thereby stabilizing sequential decision-making and improving end-to-end performance.

During inference, stage-specific policies are executed sequentially according to the pipeline. At each stage, feasible actions are selected according to the learned policy and operational constraints, producing coordinated task assignments and schedules that satisfy system requirements while reflecting the combinatorial patterns captured during training. The soft coupling mechanism, combined with trajectory augmentation, ensures that the policy is robust to variations in order arrivals, tote availability, and robot states, reducing the accumulation of errors across the sequential decision pipeline.

\subsubsection{Multi-Agent Coordination in OLSF-TRS}

To achieve scalable and robust decision making in large scale stochastic warehouse environments, the proposed framework is extended using a multi agent reinforcement learning paradigm based on MAPPO. The original sequential decision pipeline is decomposed into three cooperative agents representing orders, totes, and robots. This decomposition enables parallel and coordinated control over heterogeneous system components. Each agent $i$ receives a local observation $o_i$, which is encoded by a shared feature extractor to produce a latent representation $h_i$. To explicitly distinguish heterogeneous agent roles, an agent type embedding $e_i$ is incorporated, yielding
\begin{equation}
\tilde{h}_i = h_i + e_i.
\end{equation}
where $\tilde{h}_i$ denotes the role-aware latent representation for agent $i$.

Inter agent information is then aggregated through a multi head attention mechanism, resulting in a context aware representation
\begin{equation}
z_i = \sum_j \alpha_{ij} v_j,
\end{equation}
where $\alpha_{ij}$ is the attention weight capturing the relevance of agent $j$ to agent $i$, and $v_j$ is the value vector associated with agent $j$. The context-aware representation $z_i$ is then mapped to an agent-specific policy
\begin{equation}
\pi_i(a_i \mid o_i) = \mathrm{PolicyHead}_i(z_i),
\end{equation}
where $a_i$ denotes the action selected by agent $i$, and $\pi_i(a_i \mid o_i)$ is the corresponding policy probability.

A centralized critic provides a globally consistent learning signal by estimating the expected system-level value with respect to a unified operational objective $Z_{\mathrm{Final}}$, which represents the total number of tote movements required to complete all orders in the warehouse. Specifically, $Z_{\mathrm{Final}}$ is computed as
\begin{equation}
Z_{\mathrm{Final}} = \sum (Z_{\mathrm{Retrievals}}+ Z_{\mathrm{Returns}}),
\end{equation}
where $Z_{\mathrm{Retrievals}}$ denotes the number of totes retrieved from storage to fulfill orders, and $Z_{\mathrm{Returns}}$ denotes the number of totes returned from workstations to storage. Minimizing $Z_{\mathrm{Final}}$ directly reduces the overall tote handling workload, thereby improving operational efficiency.

The critic is implemented as a recurrent neural network (RNN) to capture temporal dependencies in warehouse operations. It consists of three main components: a spatial feature encoder, implemented as a multi-layer perceptron (MLP) with ReLU activations to transform high-dimensional global states into dense features; a temporal recurrent layer, implemented as a gated recurrent unit (GRU), which maintains hidden states to encode historical scheduling information; and a value regression head that maps the recurrent features to a scalar representing the expected cumulative reward.

Minimizing $Z_{\mathrm{Final}}$ requires tight coupling between logical batching decisions and physical execution. The framework assigns orders to totes to maximize the utility of each retrieval, ensuring that each retrieved tote satisfies as many SKU demands as possible before being returned to storage. The Robot Agent further supports this objective by minimizing makespan, which accelerates storage turnover and allows previously retrieved totes to be reused for overlapping tasks, thereby reducing redundant retrievals.

Unlike conventional heuristics that decouple batching and scheduling, the centralized critic in MAPPO treats $Z_{\mathrm{Final}}$ as a global coordination signal. By penalizing excessive tote movements, the framework encourages the Order Agent to align batching decisions with the real-time execution state of the Robot Agent and the availability of SKUs. This coordination mitigates the redundant retrieval phenomenon commonly observed in greedy strategies and enables MAPPO to learn joint policies that improve storage throughput and reduce operational uncertainty under high concurrency.

The resource-based objective inherently reflects the efficiency of the Order and Tote Agents in terms of batching quality and resource utilization, which is further supported by the Robot Agent through efficient execution. By reducing makespan, the Robot Agent increases the circulation speed of totes, lowering the cumulative need for new retrieval operations and directly contributing to the minimization of $Z_{\mathrm{Final}}$.

All agent policies are optimized cooperatively to maximize a shared global return corresponding to the minimization of $Z_{\mathrm{Final}}$. The centralized critic is trained using the mean-squared error loss:
\begin{equation}
\mathcal{L}^{\mathrm{critic}}
= \mathbb{E}\big[(V(s) - V_{\mathrm{target}})^2\big],
\end{equation}
where the target value $V_{\mathrm{target}}$ is computed via Generalized Advantage Estimation (GAE)~\citep{schulman2015high}. 

The global reward signal at time $t$ is derived from the temporal change in the system objective:
\begin{equation}
R_{\mathrm{Global}}(s_t, \mathbf{a}t)
= - \Delta Z{\mathrm{Final}}
= Z_{\mathrm{Final}}(s_t) - Z_{\mathrm{Final}}(s_{t+1}),
\end{equation}
where $\mathbf{a}_t = (a_1, a_2, a_3)$ denotes the joint action of all agents at step $t$. This ensures that each joint action is evaluated based on its immediate contribution to reducing total tote movements.

In addition to the shared global reward, each agent receives a shaped local reward $R_i$ to guide exploration and improve learning efficiency. This reward combines an intrinsic objective $f_i(s, a_i)$, inherited from the corresponding BQ-NCO pretraining module, with explicit inter-agent coordination terms:
\begin{equation}
R_i
= f_i(s, a_i)
+ \sum_{j \neq i} \lambda_{i,j} \cdot \mathrm{Impact}_{i,j}(a_i, a_j),
\end{equation}
where $f_i(s, a_i)$ is the intrinsic objective inherited from BQ-NCO pretraining, $\lambda{i,j}$ is a tunable weight for inter-agent coordination, and $\mathrm{Impact}{i,j}(a_i, a_j)$ quantifies how agent $i$’s action affects agent $j$ with respect to system efficiency (e.g., robot travel cost, order tardiness). This mechanism ensures that agents consider both local objectives and the impact on others, indirectly supporting the minimization of $Z{\mathrm{Final}}$.

Through this hierarchical integration of BQ-NCO pretraining and MAPPO based multi agent learning, the proposed framework enables scalable, coordinated, and real time decision making across heterogeneous warehouse entities. At the same time, the centralized critic enforces alignment between individual agent behaviors and the final system level operational objective.

\subsection{Baseline Methods}
To rigorously assess the performance of the proposed OLSF-TRS, we compare it against a comprehensive set of baseline methods spanning heuristic, logic-driven, and decentralized reinforcement learning approaches. These baselines include:

\begin{itemize}
        \item Collaborative SKU-Group Heuristic (C-SGH): a hybrid heuristic that balances physical routing and logical batching using a maximum-weight bipartite matching formulation.
    \item Independent PPO (IPPO): ablation baseline where each agent learns independently without a centralized critic.
    \item Robot-first Resource Routing (R$^3$) and Group-based Greedy Gathering (G$^3$): logic extremes emphasizing robot-centric and order-centric strategies, respectively.
\end{itemize}

A full description of each baseline, including algorithmic mechanisms and expected performance characteristics, is provided in Appendix~\ref{appendix:baselines}. This set of baselines enables comprehensive benchmarking of both resource efficiency ($Z_{\mathrm{Final}}$) and system total steps.

\section{Discussion}\label{sec12}

This work proposes a unified learning-based framework for large-scale warehouse order fulfillment in tote-handling robotic systems, addressing a long-standing gap between combinatorial optimization models and system-level execution under realistic operational dynamics. By decomposing the end-to-end fulfillment process into tightly coupled decision agents and integrating structured combinatorial learning with centralized multi-agent coordination, the proposed OLSF-TRS offers a scalable alternative to rule-based and monolithic optimization approaches.

A central insight of this study is that high-quality system-level performance does not require solving the full joint optimization problem explicitly. Instead, decomposing the problem into structurally meaningful subproblems and equipping each with a dedicated combinatorial policy head enables efficient local reasoning, while a centralized coordination layer resolves global coupling effects. The results from small-scale instances demonstrate that the BQ-MDP abstraction preserves the essential decision structure of individual agents, allowing the BQ-NCO policies to closely approximate optimal solutions with minimal inference latency. This confirms that bisimulation-based state compression is not merely a theoretical construct but a practical mechanism for overcoming the curse of dimensionality in sequential warehouse decision-making.

More importantly, the large-scale experiments reveal how local optimality alone becomes insufficient once system concurrency, stochastic arrivals, and resource contention dominate operational behavior. In these regimes, performance degradation is observed for independently trained policies and static heuristics, highlighting the limitations of decentralized or rule-driven control in dense robotic fleets. Specifically, decentralized methods such as IPPO suffer from the credit assignment problem: when multiple agents act simultaneously, individual reward signals become confounded by the actions of others, making it difficult for each agent to accurately attribute outcomes to its own decisions. This misattribution leads to suboptimal policy updates and, ultimately, convergence to inefficient equilibria. Static heuristics, on the other hand, lack the capacity to anticipate system-wide dynamics; they respond reactively to congestion only after it materializes, resulting in cascading delays under high-density operations.

By contrast, OLSF-TRS maintains stable performance across increasing load levels, owing to two complementary mechanisms. First, the centralized MAPPO critic observes joint system states rather than individual local observations, enabling accurate credit assignment even when reward signals are confounded by concurrent agent actions. By disentangling each agent's contribution to the shared objective, the critic provides informative gradient signals that guide policy updates toward globally coordinated behavior. Second, the value function learned by the centralized critic implicitly encodes downstream congestion patterns and resource contention dynamics. This allows agents to anticipate bottlenecks before they occur, rather than merely reacting to congestion ex post. The resulting proactive decision-making reduces unnecessary tote movements and mitigates the ripple effects of localized delays. Furthermore, the reuse of pre-trained BQ-NCO policy heads ensures that multi-agent learning focuses on coordination rather than relearning basic combinatorial structure from scratch. Since each agent already possesses a competent local policy, the MAPPO training phase concentrates sample efficiency on inter-agent dependencies, substantially reducing the exploration burden compared to end-to-end multi-agent reinforcement learning.

From an operational perspective, the observed improvements translate into tangible industrial benefits. Reductions in total tote movements directly imply lower energy consumption, reduced mechanical wear, and increased throughput stability, which are key performance indicators for large-scale automated warehouses. While the overall runtime of OLSF-TRS in simulation exceeds that of lightweight heuristics, this cost is dominated by high-fidelity environment simulation rather than policy inference. In deployment settings, where physical execution replaces simulation, the millisecond-level inference latency of OLSF-TRS enables real-time decision-making without computational bottlenecks. This decoupling of training complexity and execution efficiency aligns well with industrial requirements for reliability, responsiveness, and scalability.

The proposed framework also offers practical value for decision-makers beyond immediate operational control. Its modular structure allows warehouse designers and operators to evaluate alternative system configurations, such as station layouts, fleet sizes, or batching policies, within a unified learning environment. Rather than relying on static rules or isolated optimization tools, OLSF-TRS provides a data-driven decision support layer capable of adapting to evolving demand patterns and operational constraints. This makes the framework particularly suitable for next-generation robotic fulfillment centers, where frequent reconfiguration and continuous performance tuning are essential.

Several limitations remain. The current study focuses on homogeneous tote-handling robotic systems and abstracts away low-level motion planning and collision avoidance, which are instead captured implicitly through congestion-aware travel times. Extending the framework to heterogeneous robot fleets, integrating explicit path planning, and incorporating additional operational constraints such as battery charging and maintenance scheduling represent important directions for future work. Moreover, while the BQ-MDP abstraction is effective in structured warehouse layouts, its construction may require adaptation for highly unstructured or dynamically reconfigurable environments.

Looking ahead, the modular and hierarchical nature of OLSF-TRS opens promising avenues for further development. Integrating real-time sensory feedback would enable fully online learning and adaptation, while coupling the framework with large language models or foundation models could introduce high-level strategic reasoning, such as demand forecasting or policy explanation, into the decision loop. Together, these extensions move toward a vision of self-optimizing warehouse systems that not only execute efficiently but also reason, adapt, and evolve alongside the environments they operate in.










\appendix
\section{Detailed Descriptions of Baseline Algorithms}
\label{appendix:baselines}

\subsection{Collaborative SKU-Group Heuristic (C-SGH)}

The Collaborative SKU-Group Heuristic (C-SGH) is a high-level hybrid heuristic designed to jointly account for order batching efficiency and physical execution costs through global coordination. Rather than decomposing the problem into fully independent submodules, C-SGH adopts a rolling-horizon decision strategy in which batching logic and robot assignment are tightly coupled at each decision epoch.

At each picking station $s$, an active order pool $\mathcal{O}_{\mathrm{active}}$ with a fixed window size is maintained. For each candidate tote $t$, a batching utility function $f_{\mathrm{batch}}(o_t, s)$ is computed to quantify the degree of SKU overlap between the items stored in tote $t$ and the set of incomplete orders in $\mathcal{O}_{\mathrm{active}}$. Totes that can simultaneously satisfy a larger number of active orders are thus prioritized, implicitly performing online order batching and tote selection while reducing the total number of tote retrievals $z_{\mathrm{final}}$.

Given the filtered set of candidate tasks, the physical assignment between robots and totes is formulated as a maximum-weight bipartite matching problem. The utility of assigning robot $r$ to tote $t$ is defined as
\begin{equation}
w_{r,t} = \alpha \cdot f_{\mathrm{batch}}(o_t, s) - \beta \cdot f_{\mathrm{travel}}(r, t),
\end{equation}
where $f_{\mathrm{travel}}(r, t)$ measures the execution cost based on Euclidean or three-dimensional travel distance with vertical penalties, and $\alpha$ and $\beta$ balance batching efficiency against makespan minimization.

The resulting bipartite graph is solved using the Kuhn--Munkres (KM) algorithm, also known as the Hungarian algorithm, which finds a one-to-one assignment that maximizes the total utility $\sum w_{r,t}$~\citep{kuhn1955hungarian}. Unlike greedy or decentralized assignment rules, KM explicitly considers the global interaction among all robot--tote pairs. As a result, a locally suboptimal assignment for one robot may be selected if it enables a significantly higher overall system utility, thereby resolving conflicts and improving system-level performance.

It is worth noting that C-SGH does not explicitly formulate order batching or tote-to-order assignment as standalone optimization problems. Instead, these decisions are implicitly embedded in the rolling batching window and the SKU-affinity-driven tote selection mechanism. This design allows C-SGH to achieve high-quality coordinated decisions while remaining computationally tractable and responsive in dynamic warehouse environments.

\subsection{Independent PPO (IPPO)}
IPPO is a decentralized reinforcement learning baseline in which each agent learns independently based solely on local observations. This baseline highlights the benefits of the centralized training with shared critic in MAPPO. While agents may perform well individually, coordination is limited, leading to suboptimal system-level performance.
IPPO is a decentralized reinforcement learning baseline in which each agent learns independently based solely on local observations, following the independent learning paradigm commonly adopted in multi-agent reinforcement learning~\citep{tan1993multi}.

\subsection{Robot-first Resource Routing (R$^3$)}
R$^3$ is a robot-centric heuristic emphasizing physical execution efficiency:
\begin{enumerate}
    \item Generate robot paths based on current positions and tote availability.
    \item Treat robot idle time and proximity as key resources.
    \item Adjust order batching to fit preplanned robot paths.
\end{enumerate}
\textbf{Characteristics:} Short makespan due to efficient robot routing; potentially higher total tote retrievals ($Z_{\mathrm{Final}}$) because batching may be suboptimal.

\subsection{Group-based Greedy Gathering (G$^3$)}
G$^3$ is an order-centric batching heuristic:
\begin{enumerate}
    \item Identify order groups with maximal SKU overlap.
    \item Greedily select batches that satisfy the most orders per tote.
    \item Plan robot paths to execute the preformed batches.
\end{enumerate}
\textbf{Characteristics:} Low $Z_{\mathrm{Final}}$ due to high batching efficiency; potentially longer makespan as robots may wait or travel further to complete grouped orders.

\bibliographystyle{elsarticle-harv} 
\bibliography{cas-refs}

\end{document}